\documentclass[letterpaper, 10 pt, conference]{ieeeconf}  %

\IEEEoverridecommandlockouts                              %

\usepackage{amsmath}
\usepackage{amssymb}
\usepackage{mathtools}
\usepackage{booktabs}
\usepackage[]{graphicx}
\usepackage[font=footnotesize]{caption}
\usepackage[font=footnotesize]{subcaption}
\usepackage{gensymb}
\usepackage{lipsum}
\usepackage{float}
\usepackage{color}
\usepackage{caption}
\usepackage{threeparttable}
\usepackage{todonotes}
\usepackage[ruled]{algorithm2e}
\DontPrintSemicolon
\usepackage{siunitx}
\usepackage{cite}
\usepackage{setspace}
\usepackage{tablefootnote}

\newcommand\MYhyperrefoptions{bookmarks=true,bookmarksnumbered=true,
pdfpagemode={UseOutlines},plainpages=false,pdfpagelabels=true,
colorlinks=true,citecolor={black},
pdftitle={Redesigning SLAM for Arbitrary Multi-Camera Systems},%
pdfsubject={Computer Vision, Robotics},%
pdfauthor={J. Kuo, M. Muglikar, Z. Zhang, D. Scaramuzza},%
pdfkeywords={SLAM, Multi-Camera}}%

\makeatletter
\let\NAT@parse\undefined
\makeatother
\usepackage[\MYhyperrefoptions, pdftex, pagebackref=true,breaklinks=true,letterpaper=true,colorlinks,bookmarks=false]{hyperref} 
\definecolor{somegray}{rgb}{0.5, 0.5, 0.5}
\newcommand{\darkgrayed}[1]{\textcolor{somegray}{#1}}
\makeatletter
\newcommand*\titleheader[1]{\gdef\@titleheader{#1}}
\AtBeginDocument{%
  \let\st@red@title\@title
  \def\@title{%
    \vskip-3em
    \bgroup\normalfont\large\centering\@titleheader\par\egroup
    \vskip1.5em\st@red@title}
}
\makeatother

\titleheader{\darkgrayed{This paper has been accepted for publication at the\\
IEEE International Conference on Robotics and Automation (ICRA), Paris, 2020.
\copyright IEEE}}

\newcommand{\Sec}{Section~}
\newcommand{\Fig}{Fig.~}
\newcommand{\Tab}{Table~}
\newcommand{\Alg}{Algo.~}

\newcommand{\eg}{e.g., }
\newcommand{\ie}{i.e., }
\newcommand{\etal}{\emph{et al. }}

\newcommand{\jc}[1]{{\color{black}#1}}
\newcommand{\zz}[1]{{\color{black}#1}}

\newcommand{\px}{\mathbf{u}}

\global\long\def\pt{\mathbf{{p}}}

\global\long\def\T{\mathtt{{T}}}

\global\long\def\J{\mathtt{{J}}}

\title{\LARGE \bf
Redesigning SLAM for Arbitrary Multi-Camera Systems
}

\author{Juichung Kuo,  Manasi Muglikar, Zichao Zhang, Davide Scaramuzza%
	\thanks{The authors are with the Robotics and Perception Group, Dep. of Informatics, University of Zurich, and Dep. of Neuroinformatics, University of Zurich and ETH Zurich, Switzerland--- \url{http://rpg.ifi.uzh.ch.}
	\zz{This research was supported by the National Centre of Competence in Research (NCCR) Robotics, through the Swiss National Science Foundation, the SNSF-ERC Starting Grant and Sony R\&D Center Europe.}}%
}

\begin{document}

\maketitle
\thispagestyle{empty}
\pagestyle{empty}

\begin{abstract}
Adding more cameras to SLAM systems improves robustness and accuracy but complicates the design of the visual front-end significantly.
Thus, most systems in the literature are tailored for specific camera configurations.
In this work, we aim at an adaptive SLAM system that works for arbitrary multi-camera setups.
To this end, we revisit several common building blocks in visual SLAM.
In particular, we propose an adaptive initialization scheme, a sensor-agnostic, information-theoretic keyframe selection algorithm, and a scalable voxel-based map.
These techniques make little assumption about the actual camera setups and prefer theoretically grounded methods over heuristics.
We adapt a state-of-the-art visual-inertial odometry with these modifications, and experimental results show that the modified pipeline can adapt to a wide range of camera setups (\eg 2 to 6 cameras in one experiment) without the need of sensor-specific modifications or tuning.

\end{abstract}

\section*{Supplementary Material}
Video: \href{https://youtu.be/JGL4H93BiNw}{https://youtu.be/JGL4H93BiNw}

\section{Introduction}\label{sec:intro}

As an important building block in robotics, visual(-inertial) odometry (VO/VIO), or more general, simultaneous localization and mapping (SLAM) has received high research interest.
Modern SLAM systems are able to estimate the local motion accurately as well as build a consistent map for other applications.
One of the remaining challenges for vision-based systems is the lack of robustness in challenging environments, such as high dynamic range (HDR) and motion blur~\cite{Cadena16tro}.
Among different approaches that have been explored for better robustness (\eg \cite{Zhang17icra} \cite{Rosinol18ral}), adding more cameras in SLAM systems proves to be effective and is already exploited in successful commercial products, such as Oculus Quest \cite{oculus_quest} and Skydio \cite{skydio_r1}.

As the workhorse for modern (keyframe-based) SLAM systems, bundle adjustment (BA) like nonlinear optimization naturally generalizes to multiple sensors, including visual-inertial and multi-camera systems, as long as the measurement process is modeled correctly.
On the other hand, the design of the so-called front-ends is much less theoretically grounded.
Many details, such as initialization, keyframe selection, and map management, are designed heuristically.
Moreover, such designs are often tailored to specific sensor setups, and it is not clear to what extent they can be applied to more general sensor configurations.
For example, one popular method for selecting keyframes is to consider commonly visible features in the current frame with respect to the last keyframe.
While this works well for monocular setups or stereo pairs with highly overlapping field-of-views (FoV), it quickly becomes complicated as more cameras are added, as different cameras may have drastically different view conditions (\eg the number of features).

\begin{figure}
    \centering
    \includegraphics[width=0.99\linewidth]{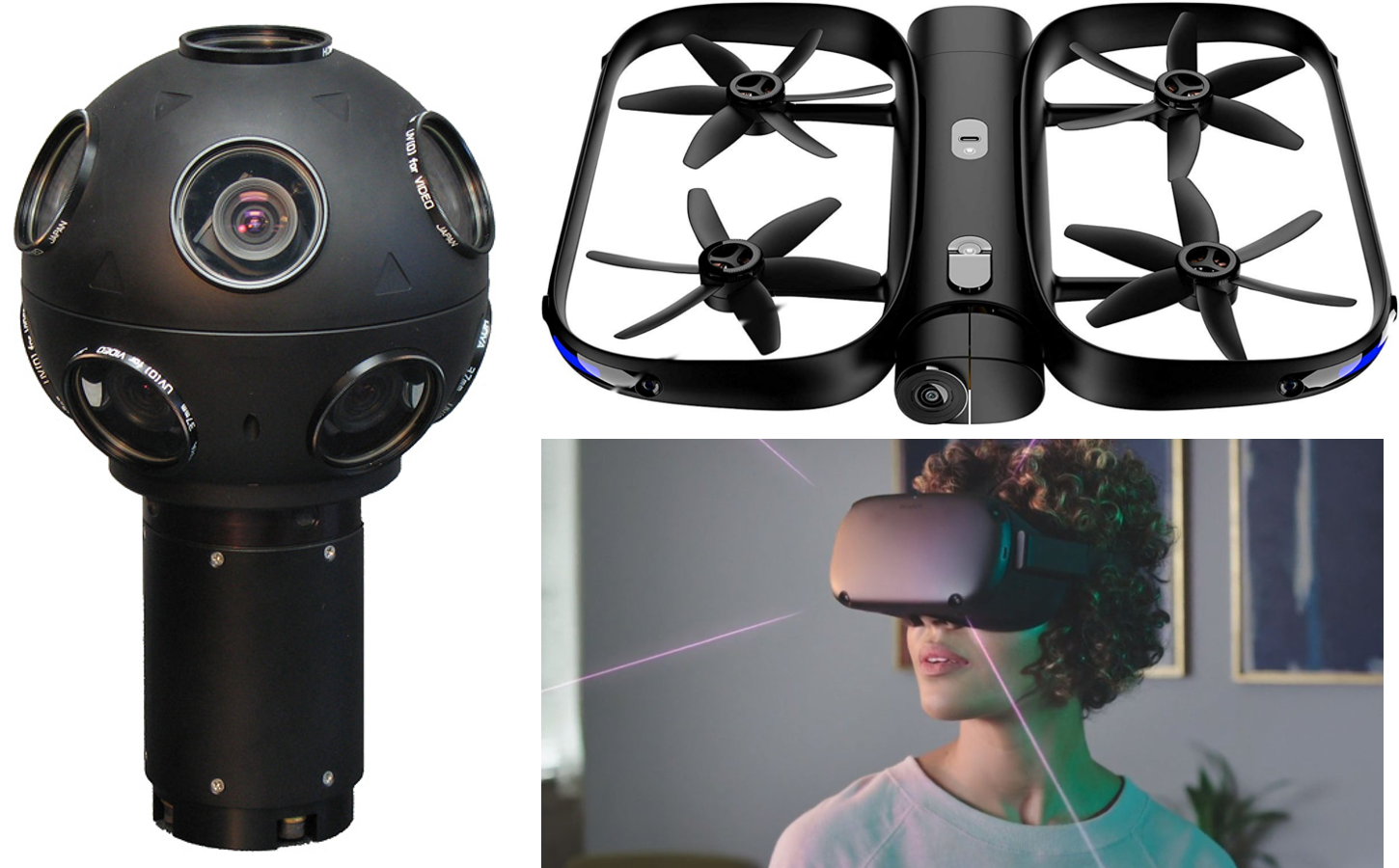}
    \caption{
    \zz{
    Multi-camera systems achieve superior performance in perception algorithms and are widely used in real-world applications, such as omnidirectional mapping \cite{google_street_view}, autonomous drones \cite{skydio_r1}, and VR headsets\cite{oculus_quest}.
    To facilitate the use of such systems in SLAM, we propose several generic designs that adapt to arbitray multi-camera systems automatically.
    }
    }
    \label{fig:eye_catcher}
    \vspace{-4ex}
\end{figure}

To remove the dependence on sensor-specific assumptions and heuristics, we resort to adaptive and more principled solutions.
First, instead of using hard-coded rules, we propose an adaptive initialization scheme that analyzes the geometric relation among all cameras and selects the most suitable initialization method online.
Second, instead of engineering heuristics, we choose to characterize the uncertainty of the current pose estimate with respect to the local map using the information from all cameras, and use it as an indicator of the need for a new keyframe.
Third, instead of relying on the covisiblity graph, we organize all the landmarks in a voxel grid and sample the camera frustums via an efficient voxel hashing algorithm, which directly gives the landmarks within the FoVs of the cameras.
These methods generalize well to arbitrary camera setups without compromising the performance for more standard configurations (\eg stereo).

\noindent
\textbf{Contributions}: To summarize, the contribution of this work is an adaptive design for general multi-camera VO/VIO/SLAM systems, including
\begin{itemize}
    \item an adaptive initialization scheme,
    \item a sensor-agnostic, information-theoretic keyframe selection algorithm,
    \item a scalable, voxel-based map management method.
\end{itemize} 
Since the proposed method is not limited to specific implementations or sensing modalities, we will use the term SLAM in general for the rest of the paper.

The paper is structured as follows.
In \Sec\ref{sec:related_work}, we review the common methods for initialization, keyframe selection, and map management in visual SLAM.
In \Sec\ref{sec:initialization}, we describe our adaptive initialization process based on overlap check.
In \Sec\ref{sec:entropy_kf}, we detail our entropy-based keyframe selection algorithm.
In \Sec\ref{sec:voxel_map}, we introduce our voxel-based map representation for visible points retrieval.
Finally, we apply our method to a state-of-the-art VIO system and present the experimental results in \Sec\ref{sec:experiments} and conclude our work in \Sec\ref{sec:conclusion}.

\section{Related Work}\label{sec:related_work}

\subsection{Initialization}\label{ssec:bg_init}
The initialization in SLAM typically incorporates assumptions that are specific to camera configurations.
For monocular systems, homography and 5-point relative pose algorithm from \cite{Nister04pami} are popular ways to obtain the poses of the first two keyframes and the initial map (\eg \cite{MurArtal15tro}), which usually requires the camera to undergo certain motion, such as strong translation and no pure rotation.
In contrast, stereo cameras can recover the depth information and initialize the map directly \cite{MurArtal17tro, Forster17troSVO}.
In multi-camera setups, there could be various ways of combining different cameras depending on the extrinsic parameters.
For example, MCPTAM \cite{Harmat2012ICIRA} initializes the monocular cameras individually with a known target.
The pipeline in Liu \etal \cite{Liu2018IROS} performs initialization with stereo matching from predefined stereo pairs.
While the possible ways for initialization inevitably depend on sensor configurations, we would like a system to be able to select the proper method automatically, instead of hard-coded rules.

\subsection{Keyframe Selection}\label{ssec:bg_kf}
It is common to maintain a fixed number of keyframes in the front-end as the local map, against which new frames are localized.
Hence, the selection of keyframes is  crucial for the performance of SLAM systems.
In general, the keyframe selection criteria can be categorized into heuristic-based methods and information-theoretic methods.

\subsubsection{Heuristics-based methods}\label{sssec:bg_heuristics}
In many SLAM systems, the keyframe selection criteria are the combinations of different heuristics. We list the most common ones below.

\noindent
\textbf{Camera motion:}
In ORB-SLAM \cite{MurArtal15tro}, one of the criteria is to check whether the current frame is a certain number of frames away from the last keyframe.
Similarly, DSO \cite{Engel17pami} and SVO \cite{Forster17troSVO} select a new keyframe if the current pose is away from the last keyframe by a certain amount of motion.

\noindent
\textbf{Number of tracked features:}
A new keyframe is selected if the number or the percentage of tracked features in the current frame falls below a certain threshold.
However, the specific threshold usually varies greatly between different scenarios.
This criterion in used in \cite{MurArtal15tro}, \cite{Forster17troSVO}, \cite{Harmat2012ICIRA}, and \cite{Qin18tro}.

\noindent
\textbf{Optical flow:}
The Euclidean norm between the corresponding features from the current frame and the last keyframe.
This criterion, for example, is used in  \cite{Engel17pami} and \cite{Liu2018IROS}. 

\noindent
\textbf{Brightness change:}
For direct methods, changes in image brightness caused by camera exposure time and lighting condition makes the tracking against old keyframes difficult.
Hence, \cite{Engel17pami} also uses the relative brightness as a criterion.

Using the combination of different heuristics usually relies on certain assumptions of the sensor configurations and scenes, which makes parameter tuning as well as the application to general multi-camera setups complicated.

\subsubsection{Information-theoretic methods}\label{sssec:bg_info}
These methods rely on more principled metrics and are less common in literature.
Das \etal \cite{Das2015IROS} chose the keyframes to be included in the BA.
Their method favors the frames that bring the most entropy reduction in the map points and essentially selects the most informative keyframes for BA.
The criterion from DVO \cite{Kerl13iros} is the most related to ours:
it selects keyframes based on an entropy ratio that reflects the uncertainty of the camera pose with respect to the last keyframe.
Our method follows a similar idea, but considers all the current keyframes, which reflects the information contained in the entire local map.

\subsection{Map Management and Query}\label{ssec:bg_mm}
To estimate the pose of newly coming frames, the front-end usually needs to find the 2D-3D correspondences between the observations in the new images and the map.
A common method is the covisibility check: only search for the matches of the 3D points in the keyframes that reproject onto the new images, such as in \cite{Klein07ismar, Forster14icra, MurArtal15tro, Engel17pami}.
As more cameras are added, the complexity of the covisiblity check increases quadratically, and keyframes from cameras with large common FoVs introduce high redundancy.
For example, for stereo pairs with highly overlapping FoVs, it is usually sufficient to keep one of the two frames as keyframes. Obviously, it is not clear how this strategy can generalize to arbitrary camera configurations.
To the best of of our knowledge, there is no previous study about efficiently querying map points in a general multi-camera setup. 
\begin{figure}[t]
     \centering
     \includegraphics[width=0.25\textwidth]{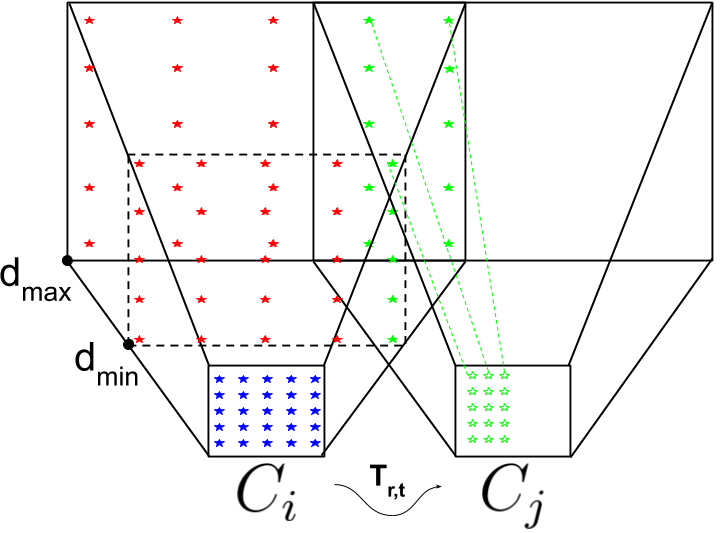}
     \captionsetup{aboveskip=5pt}
     \caption{An illustration of the stereo overlapping check between two cameras, $C_{i}$ and $C_{j}$.
     The blue stars are the sampled points on the image plane of camera $i$.
     The green stars are the 3D points that are successfully projected to camera $j$, and the red ones are the points that fall out of the image plane.
     }
     \label{img:overlap_check}
     \vspace{-4ex}
\end{figure}

\section{Adaptive Initialization}\label{sec:initialization}

Our initialization method has no hard-coded assumptions regarding the camera configuration.
For any multi-camera setups with known intrinsic and extrinsic calibrations, it is able to select the proper initialization method accordingly, without the need to change the algorithm settings manually.
Specifically, it utilizes an overlapping check between the camera frustums to identify all the possible stereo camera pairs.
If there exists stereo pairs, the initial 3D points are created from the stereo matching of these stereo pairs.
Otherwise, the 5-point algorithm is run on every camera as in a standard monocular setup, and the map is initialized whenever there exists a camera that triangulates the initial map successfully (\ie enough parallax, and the camera is not undergoing strong rotation).

The core part of the aforementioned initialization scheme is the overlapping check.
The overlapping checking algorithm checks all the possible pairs in a multi-camera configuration, denoted as $C_{ij}$, where $i,j \in 1 \dots n,\ i \neq j$, and $n$ is the total number of cameras in the system, and finds all possible stereo pairs.
For each pair $C_{ij}$, the algorithm is illustrated in \Fig \ref{img:overlap_check}.
We denote a 3D point in homogeneous coordinates as $(x,y,z,1)^{\top}$. 
With the camera projection function $\pi$, a 2D point $\px$ in the image plane can be back-projected to a 3D point in the camera frame for a depth value $z$ as $\pt = \pi^{-1}(\px, z)$.
We also know the corresponding relative transformation $\T_{ij}$ from the extrinsic calibration of the camera system.
In detail, the overlapping check first uniformly samples (or possibly using different sampling methods) a set of points $U_i$ on the image plane of camera $i$.
Then the points in $U_i$ are back-projected to the minimal and maximal depths $d_{min}$ and $d_{max}$ as $P_{i, max}$ and $P_{i, min}$ respectively.
These depths are specified by users and encloses the effective depth range of the initialization process.
Given the $\T_{ij}$ and the intrinsics of camera $j$, we then project the 3D points $P_{i, min}$ and $P_{i, max}$ to camera $j$ as $U_{j, min}$ and $U_{j, min}$ and check whether these points fall in the image plane of camera $j$.
The projection from $\px_i$ in $U_i$ to camera $j$ is considered successful only if both of the backprojected 3D points at $d_{min}$ and $d_{max}$ are within the image plane of camera $j$.
A pair of cameras is considered as a stereo pair if the overlapping ratio, $\frac{\text{\# of Successful Projection}}{\text{\# of Total Samples}}$, is above a user-defined threshold.

The proposed sampling-based method is generic. 
By using the camera projection/backprojection directly, we can find all stereo pairs across different types of camera models without the need to explicitly \zz{calculate the overlapping volume of possibly very different frustums (e.g., between pinhole and fisheye cameras), which can be non-trivial to compute analytically.}
Moreover, the check can be computed offline, and the valid stereo pairs be directly used at runtime. 

\begin{figure}[t]
     \centering
     \includegraphics[width=0.45\textwidth]{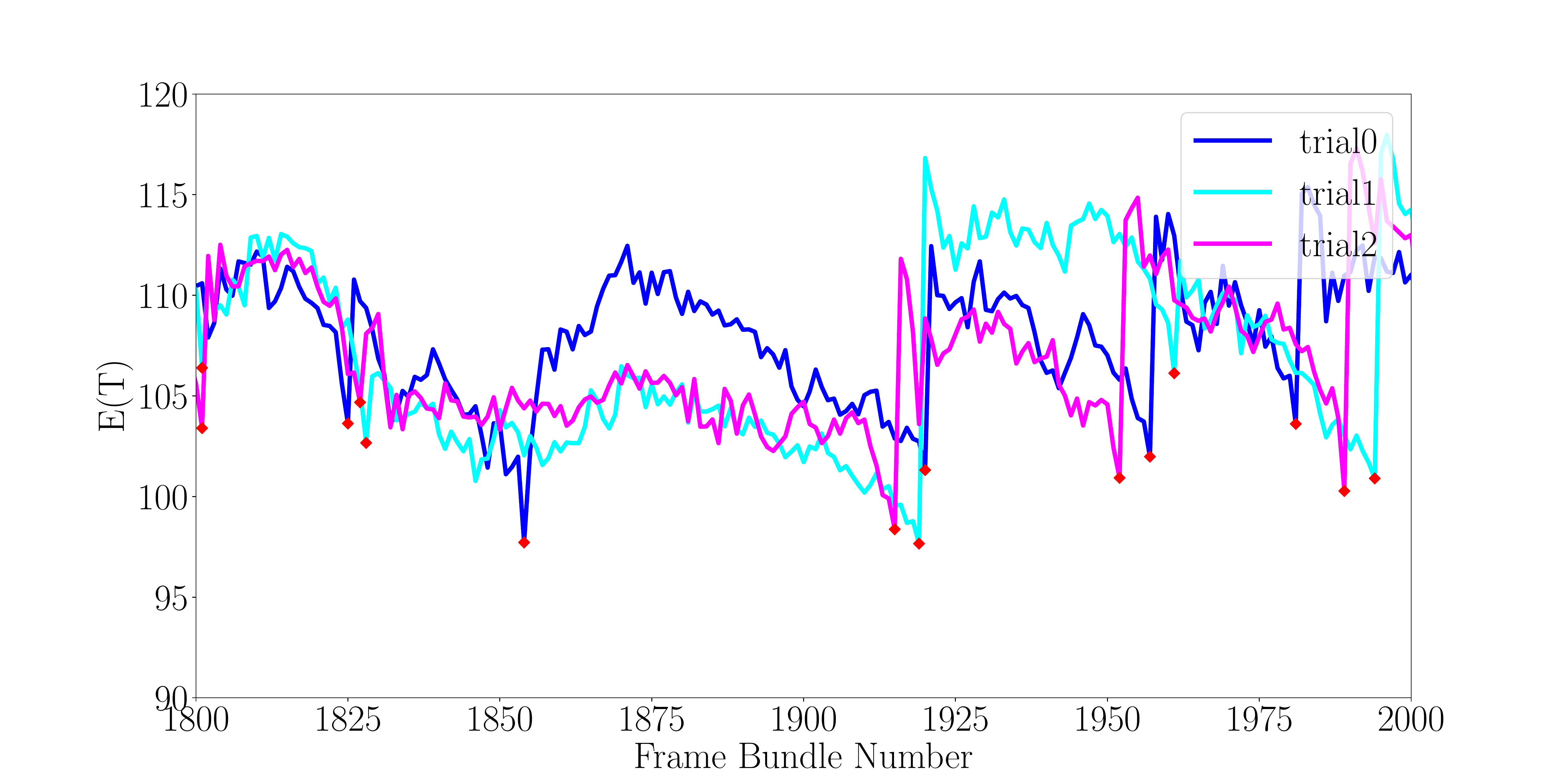}
     \captionsetup{aboveskip=5pt}
     \caption{
     Negative entropy evolution of 3 runs in EuRoC \emph{MH\textunderscore01}.
     $E(\T)$ for each run is shown in a different color,
     and the red dots indicates where a frame is selected as a keyframe.
     $E(\T)$ increases after a keyframe insertion and decreases as the sensor moves away from the map.
     }
     \label{img:pose_refinement_entropy_check}
     \vspace{-4ex}
\end{figure}

\begin{figure}[t]
     \centering
     \includegraphics[width=0.45\textwidth]{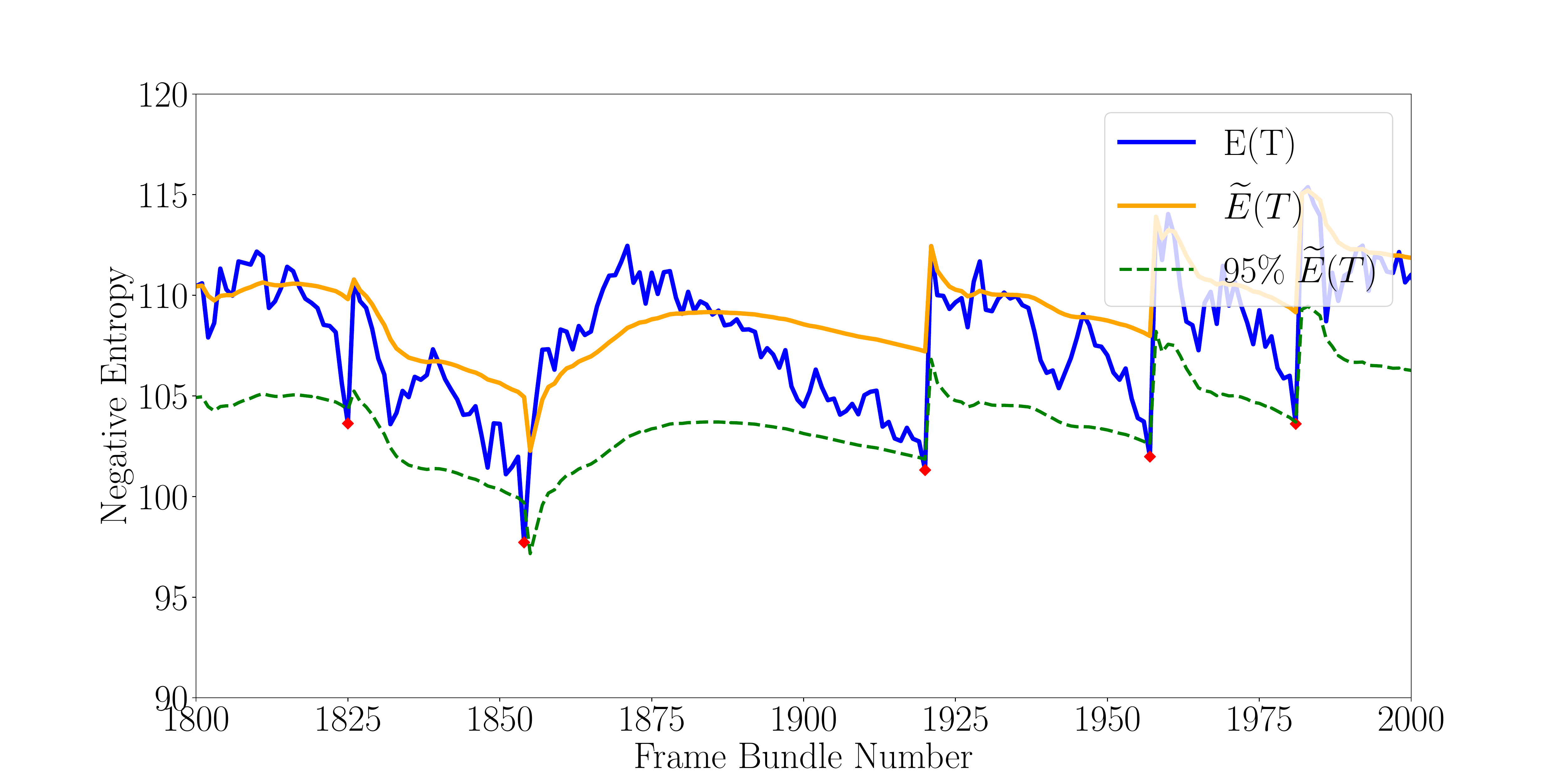}
     \captionsetup{aboveskip=5pt}
     \caption{
     Running average $\tilde{E}(\T)$ and keyframe selection.
     The running average filter (yellow) tracks the localization quality since the last keyframe.
     When the negative entropy of the current frame (blue) falls below a certain percentage of the running average (green dash), a new keyframe is selected (red dots) and the running average filter is reset. 
     }
     \label{img:running_average_filter_for_entropy}
     \vspace{-4ex}
\end{figure}

\section{Entropy-based Keyframe Selection}\label{sec:entropy_kf}

The concept of keyframe naturally generalizes to a keyframe bundle for a multi-camera setup, as in \cite{Harmat2012ICIRA}.
A keyframe bundle contains the frames from all the cameras at the same time.
In the following, we will use the terms keyframe and keyframe bundle interchangeably.
To determine when a keyframe should be added, we design an entropy-based mechanism.
In particular, the local map contains 3D points (organized as keyframes or voxels as in \Sec\ref{sec:voxel_map}) against which new frames can localize.
Intuitively, a keyframe should be selected when the current map is not sufficient for tracking, since new points will be initialized at the insertion of a keyframe.
Therefore, we select keyframes based on the uncertainty of the keyframe bundle pose with respect to the current map.
Compared with heuristics, our method is more principled, has less parameters (\emph{only 1}) and generalizes to arbitrary camera configurations.
In this section, we first provide necessary background on the uncertainties in estimation problems and then describe our keyframe selection method.

\subsection{Uncertainties Estimation in Nonlinear Least Squares}
\label{ssec:param_uncertainty}

For a parameter estimation problem of estimating $\mathbf{x}$ from measurement $\mathbf{z}$ with normally distributed noise, a common method is to cast the problem as a nonlinear least squares (NLLS) problem.
In iterative algorithms of solving NLLS problems, such as Gauss-Newton, the uncertainties of the estimated parameters can be obtained as a side product in each iterative step.
Specifically, the normal equation at step $i$ is $(\J^{\top}\Sigma_{\mathbf{z}}^{-1}\J) \delta\mathbf{x}_i = \J^{\top}\mathbf{r}(\mathbf{x}_i)$,
where $\mathbf{r}(\mathbf{x}_i)$ is the residual given the current estimate $\mathbf{x}_i$,  $\delta\mathbf{x}_i$ the optimal update, and $\J$ the Jacobian of $\mathbf{z}$ with respect to $\mathbf{x}$.
With first-order approximation, the covariance of the estimate can be obtained by backward propagating the measurement noise to the parameters, which is simply:
\begin{equation}
\mathtt{\Sigma}_{\mathbf{x}} = (\J^{\top} \mathtt{\Sigma}_{\mathbf{z}}^{-1} \J)^{-1},
\end{equation}
which is an important tool to quantify the estimation quality of NLLS solutions \cite[Chapter~5, App.~3]{Hartley03book}.
$\mathtt{I}_{\mathbf{x}} = \J^{\top}\mathtt{\Sigma}_{\mathbf{z}}^{-1}\J$ is also known as the Fisher information.

\subsection{Negative Pose Entropy in SLAM}
\label{ssec:fim_slam}
In keyframe-based SLAM, the pose of the current camera is usually obtained by solving a NLLS problem.
For example, one common method is to solve a Perspective-n-Points (PnP) problem using the Gauss-Newton method.
In this case, the Fisher information and the covariance of the camera pose can be directly obtained as
\begin{equation}
    \mathtt{I}_{\T}= \J_{\T}^{\top}\mathtt{\Sigma}_{\px}^{-1}\J_{\T},\;\mathtt{\Sigma}_\mathtt{T} = (\J_{\T}^{\top}\mathtt{\Sigma}_{\px}^{-1}\J_{\T})^{-1},
    \label{eq:pose_fim_cov}
\end{equation}
where $\px$ is the observation, and $\J_{\T}$ is Jacobian of $\px$ with respect to the camera pose $\T$.
\footnote{Technically, the Jacobian is with respect to a minimal parameterization of 6 DoF poses, which is omitted here for easy presentation.}
Note that in different NLLS problems, the Fisher information and covariance may be obtained differently (\eg marginalization in a BA setup).

As mentioned above, our goal is to use the estimate uncertainty of the current pose to indicate whether a new keyframe should be inserted.
While \eqref{eq:pose_fim_cov} provides a principled tool, it is more desirable to have a scalar metric as keyframe selection criteria.
Therefore, we utilize the concept of the differential entropy for a multivariate Gaussian distribution, which is 
$
 H(\mathbf{x}) = \frac{1}{2}m(1+\ln(2\pi))+\frac{1}{2}\ln(\begin{vmatrix}\mathtt{\Sigma}\end{vmatrix})
$
for a $m$-dimensional distribution with covariance $\mathtt{\Sigma}$.
Note that the magnitude of the entropy only depends on $\ln(\begin{vmatrix}\mathtt{\Sigma}\end{vmatrix})$.
Moreover, in the context of NLLS for pose estimation, from \eqref{eq:pose_fim_cov}, we have $\ln(|\mathtt{\Sigma_{\T}}|) = -\ln(|\mathtt{I}_{\T}|)$.
Since that the Fisher information $\mathtt{I}_{\T}$ comes for free in the process of solving NLLS problems, we can actually avoid the matrix inversion and use
\begin{equation}
    E(\mathtt{T}) \triangleq \ln(|\mathtt{I}_{\T}|)
    \label{eq:neg_entropy}
\end{equation}
to indicate how well the camera can localize in the current map.
We refer to \eqref{eq:neg_entropy} as \emph{negative entropy}.
Since \eqref{eq:pose_fim_cov} is simply the sum of individual measurements, it is straightforward to incorporate the observations from all the cameras into one single scalar \eqref{eq:neg_entropy} in an arbitrary multi-camera setup.

\setlength{\textfloatsep}{2pt}
\begin{algorithm}[t]
\scriptsize
\SetAlgoVlined
\KwIn{newest entropy value $E$(T)}
\KwResult{Returns the current running average, $\widetilde{E}(T)$}
initialization:
$n = 0,\;\widetilde{E}(T) = 0$\\
\For{\textsf{each incoming} $E(T)$}
{
    $n = n + 1$\\
    $\widetilde{E}(T) = \widetilde{E}(T) + (E(T)-\widetilde{E}(T))/n$\\
    $return\ \widetilde{E}(T)$\\ 
}
 \caption{Running average filter.}
 \label{running_average_algoritm}
\end{algorithm}

\subsection{Running Average Filter for Keyframe Selection}
\label{ssec:running average_kf_select}

Examples of the negative entropy $E(\T)$ evolution on the same dataset (\emph{MH\textunderscore01}) with our multi-camera pipeline (see \Sec\ref{sec:experiments}) are shown in \Fig\ref{img:pose_refinement_entropy_check}.
We can see that $E(\T)$ indeed reflects the localization uncertainty of the pose with respect to the current map.
After inserting a new keyframe to the map (red dots on the curves), the negative entropy increases, due to the triangulation of new points; and as the camera moves away from the last keyframe/local map, $E(\T)$ decreases until another keyframe is selected.
On the other hand, even for the same environment, the absolute value of $E(\T)$ varies from run to run.
This indicates that using an absolute threshold for $E(\T)$ as the keyframe selection criterion is not feasible.

Instead, we propose to track the negative entropy value using a running average filter (see \Alg\ref{running_average_algoritm}) in the local map  and selects a keyframe when $E(\T)$ of a frame is below certain percentage of the tracked average $\tilde{E}(\T)$.
Since we localize the camera with respect to the latest map, and the map remains the same until a new keyframe is added, $\tilde{E}(\T)$ essentially tracks the average pose estimation quality with respect to the local map up to the current time.
Note that the running average filter is reinitialized every time the map is updated with a new keyframe, since the local map changes as a new keyframe is inserted.
Moreover, we use a relative threshold with respect to the running average $\tilde{E}(\T)$ so that the selection is adaptive to different environments.
This threshold is the only parameter in our keyframe selection method, and it is intuitive to tune.
A higher value means more frequent keyframe insertion, and vice versa (see \Tab\ref{tab:euroc_kf_comparison_BA}).
An example of the running average filter is shown in \Fig \ref{img:running_average_filter_for_entropy}.

\setlength{\textfloatsep}{20.0pt plus 2.0pt minus 4.0pt} 
\section{Voxel-Map Query}
\label{sec:voxel_map}

For new incoming images, the tracking process in SLAM is responsible to find the correspondences between the observations in the new images and the 3D points in the map.
For monocular and stereo setup, this can be efficiently done by searching only for matches of the points in the keyframes that overlap with the new frames.
For a general multi-camera setup, since keyframes from different cameras can have high overlap, this method can introduce considerable redundancy.
Therefore, we organize the map points in a voxel grid, and directly sample the camera frustums for possible 3D points to match, \zz{as proposed in \cite{Muglikar2020ICRA}}.

\noindent
\textbf{Map representation:}
Our voxel-map is a hash table using the voxel hashing technique described in \cite{Niessner13tog}.
Each entry in the hash table is a voxel of a user-defined size at a certain position, and it contains the 3D points (from SLAM pipeline) that fall in this voxel.
The voxels in the hash table are accessed via a hashing function on the integer world coordinates.
Therefore, to get the 3D points around a location of interest, we can directly get the corresponding voxel in constant time.
In addition, the map only allocates voxels where there are 3D points and does not store empty voxels.
The voxel hash table is synchronized with the map points in the SLAM pipeline.

\noindent
\textbf{Map query:} %
To get the map points to match for a multi-camera system, we sample a fixed number of points in the camera frustums and find corresponding voxels in the voxel-map.
The points inside these voxels are then used to match the observations in the new images.
In this way, it is guaranteed that \emph{all and only} the 3D points within the FoVs of all cameras are retrieved from the map.
Moreover, we avoid the process of checking overlapping keyframes from different cameras, which may have many points in common and introduce redundant computation.

\zz{Note that we only use voxel-map for querying visible landmarks. Keyframes are still selected for triangulation and potentially bundle adjustment}. 
\begin{figure}[t]
    \centering
    \includegraphics[width=0.7\linewidth]{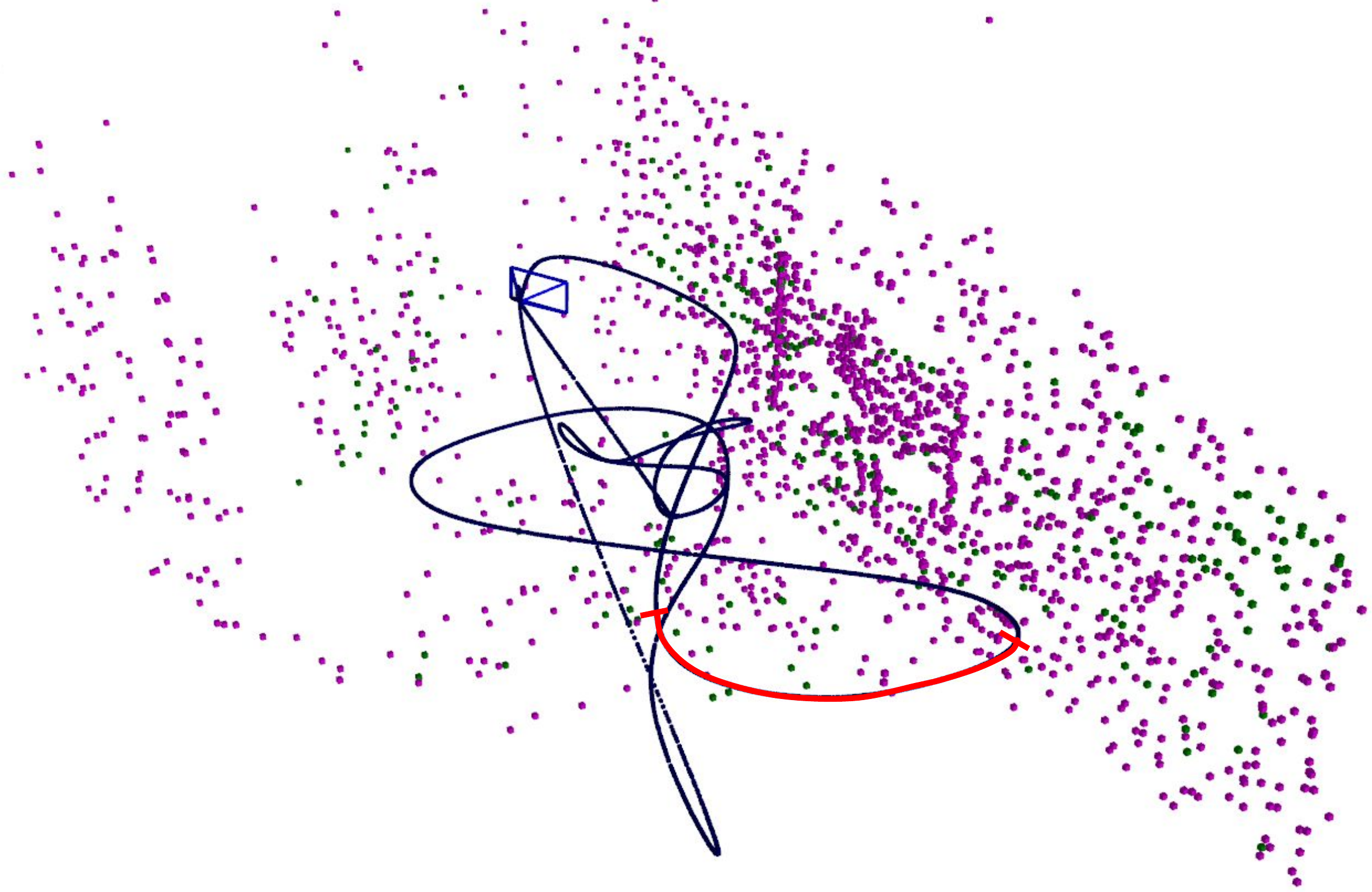}
    \caption{
    Simulated figure 8 trajectory in the simulation environment.
    The trajectory was estimated by running the adapted VIO pipeline with 5 cameras.
    The segment where the monocular setup lost track is marked in red.
    The magenta dots are the tracked landmarks by SLAM systems.}
    \label{fig:sim_traj}
    \vspace{-4ex}
\end{figure}

\section{Experiments}\label{sec:experiments}
To validate the proposed method, we applied the aforementioned adaptations to a state-of-the-art keyframe-based visual-inertial odometry pipeline that consists of an efficient visual front-end \cite{Forster17troSVO} and an optimization-based backend similar to \cite{Leutenegger15ijrr}.
We performed experiments on both simulated and real-world data.
In simulation, we verified the robustness and analyzed several properties of the pipeline with different multi-camera configurations.
For real-world data, we first tested the stereo setup with the EuRoC dataset \cite{Burri15ijrr} to show that the proposed method performs on par with standard methods but is much easier to tune.
We then tested the multi-camera setup with the AutoVision dataset \cite{autovision_dataset}.
For quantitative evaluation of accuracy, we follow the evaluation protocol in \cite{Zhang18iros}.
We repeated the experiment on each sequence for 5 runs using the same setting unless specified otherwise.
In each of the experiment, we kept the parameters \emph{the same} for different camera configurations.

\vspace{-1ex}
\subsection{Simulation Experiment}\label{ssec:exp_simulation}

We tested the pipeline on a drone with various camera configurations:
2 cameras (a front mono; a side mono), 3 cameras (a front stereo; a side mono), 4 cameras (a front stereo; a side stereo), and 5 cameras (a front stereo; a side stereo; a down mono).
\zz{We refer the reader to the accompanying video for the visualization of our setup.}
We set the drone to fly a figure 8 trajectory in the environment (\Fig\ref{fig:sim_traj}).
Note that a monocular setup from either the front or side stereo failed when the drone went around a textureless pillar (marked in red in \Fig\ref{fig:sim_traj}), and the corresponding quantitative results are omitted.
Next, we analyzed the accuracy and timing, and the performance of voxel-map and keyframes.

\begin{figure}[]
     \centering
     \includegraphics[width=0.4\textwidth]{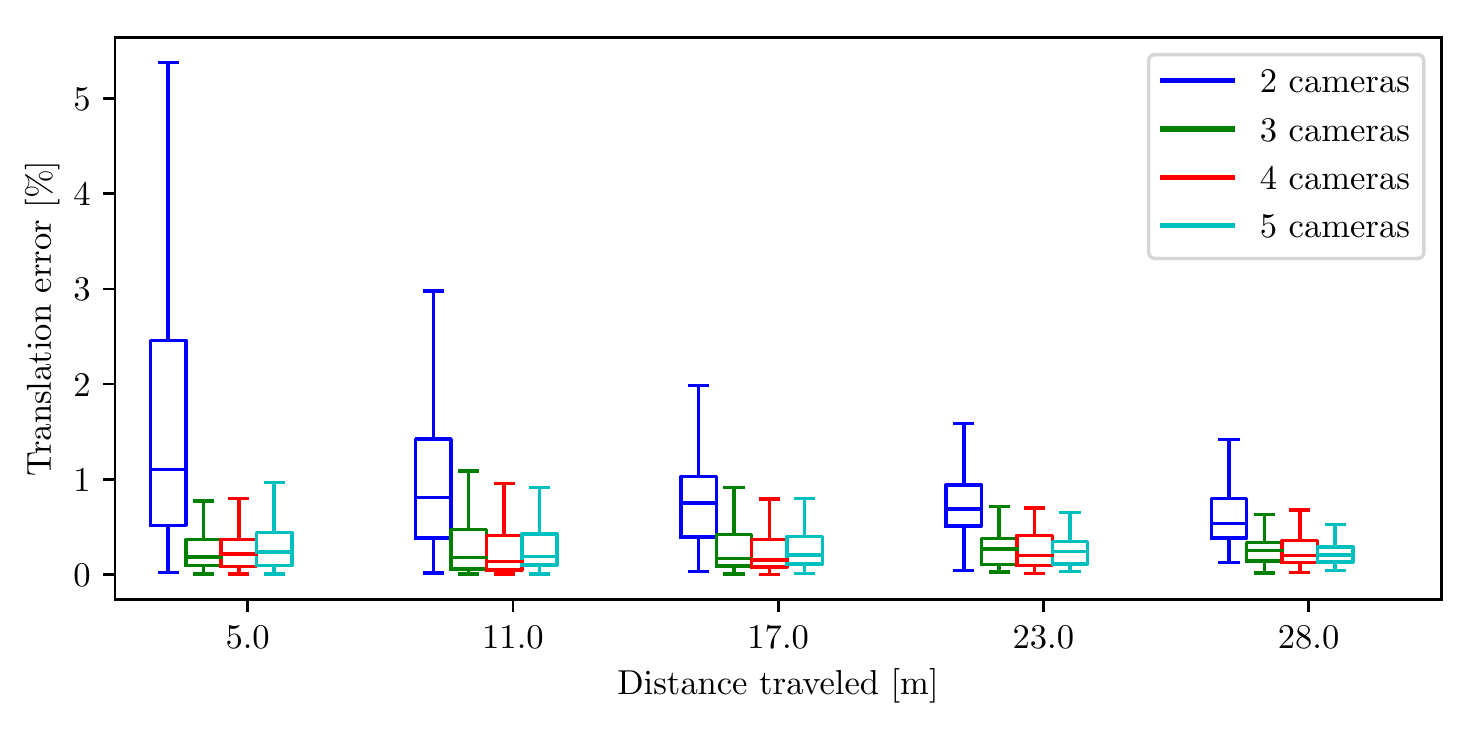}
     \caption{Overall relative translation error in simulation for 5 runs.}
     \vspace{-2ex}
     \label{img:sim_traj_error_comparison}
\end{figure}

\begin{figure}[t!]
    \centering
    \includegraphics[width=0.5\textwidth]{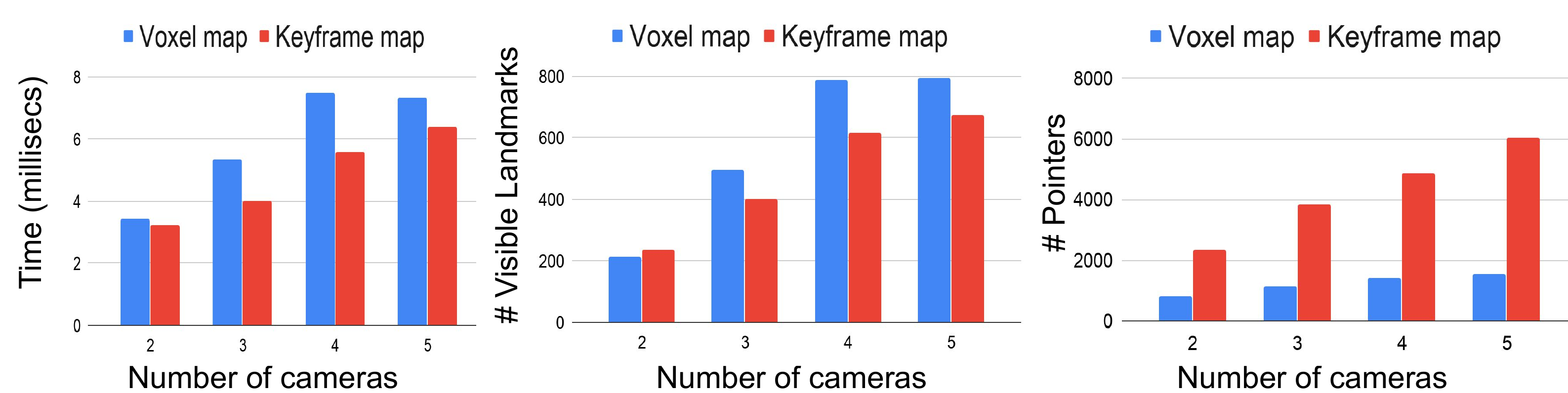}
    \caption{
    Comparison of the proposed voxel-map with standard keyframes for different camera configurations (2 to 5 cameras).
    \textbf{Left}: total time for the front-end in VIO.
    \textbf{Middle}: retrieved landmarks for matching from the map.
    \textbf{Right}: number of references/pointers to landmark positions.
    }
    \label{img:simulation_data}
    \vspace{-4ex}
\end{figure}

\noindent
\textbf{Accuracy:}
The relative pose error of different camera configurations is shown in \Fig \ref{img:sim_traj_error_comparison}. Adding more cameras to the system improved the trajectory estimation accuracy, but the improvement became marginal after the 3-camera configuration.
This is because adding the third camera formed a stereo pair (front stereo) compared with the 2-camera configuration, which made direct triangulation possible.
\noindent
\textbf{Timing:}
The total front-end time for different configurations is shown in \Fig\ref{img:simulation_data} (left).
As we increased the number of cameras in the configuration, we observe an increase in the total processing time of the front-end.
The increase in time is not as significant between the 4 and 5 camera configurations, as the 5th camera (downlooking) did not produce as many landmarks as the other cameras.

\noindent
\textbf{Voxel-map vs. Keyframes:}
In general, the voxel-map method retrieved more landmarks (\Fig\ref{img:simulation_data} middle) than the keyframe based method, \zz{because some of the visible landmarks in the current frame may not be stored in nearby keyframes. However, the front-end consumed more time in our experiment, and we assume that it can be further reduced by optimizing our voxel-map implementation}.
In terms of memory footprint (\Fig\ref{img:simulation_data} right), the voxel-map increased much slower than keyframes.
The reason is that the keyframe-based map stores landmark observations in each keyframe.
For a multi-camera setup with large FoV overlap, it is very likely different cameras observe the same landmarks, resulting in redundant copies in keyframes.
In contrast, the voxel-map stores the references only once.

\vspace{-1ex}
\subsection{Real-world Experiment}\label{ssec:exp_real}

\begin{figure*}[t!]
    \centering
    \includegraphics[width=\textwidth]{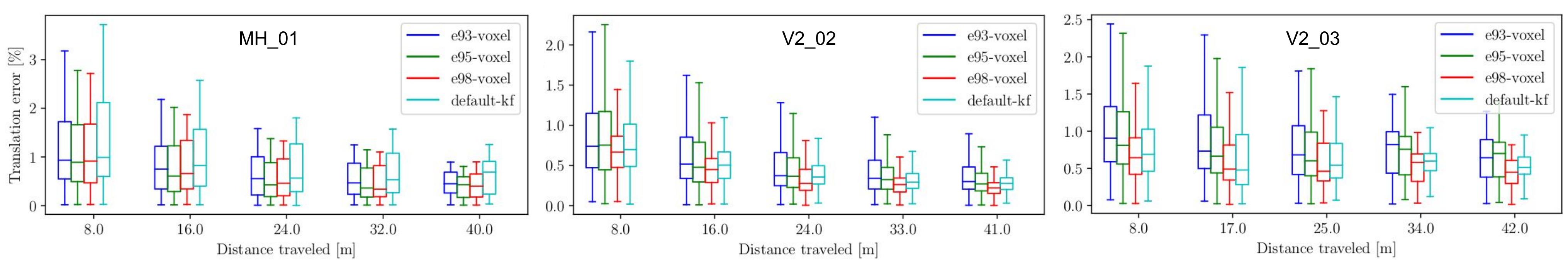}
    \caption{Relative translation error percentages from the EuRoC dataset with BA.}
    \label{fig:euroc_rel_traj_error}
    \vspace{-2ex}
\end{figure*}

\begin{table*}[t]
    \scriptsize
    \centering
    \caption{Median RMSE (meter) on EuRoC dataset over 5 runs. Lowest error highlighted in \textbf{bold}.}
    \begin{tabular}{c|c c c c c c c c c c c}
    \hline
    \textbf{Algorithm}  &      \textbf{MH\_01} & \textbf{MH\_02} & \textbf{MH\_03} & \textbf{MH\_04} & \textbf{MH\_05} & \textbf{V1\_01} & \textbf{V1\_02} & \textbf{V1\_03} & \textbf{V2\_01} & \textbf{V2\_02} &  \textbf{V2\_03} \\ \hline
    default-kf & 0.140 & 0.078 & 0.091 & \textbf{0.119} & 0.330 & 0.042 & 0.070 & 0.047 & 0.056 & 0.066 & 0.127 \\
    e93-voxel & 0.104 & 0.390 & 0.107 & 0.177 & 0.262 &\textbf{0.038} & \textbf{0.036} & 0.043 & 0.080 & 0.103 &  0.169 \\
    e95-voxel & \textbf{0.078} & 0.084 & 0.093 & 0.182 & 0.237 & 0.040 & 0.047 & 0.049 & 0.056 & 0.087 & 0.171 \\
    e98-voxel & 0.095 & \textbf{0.074} & \textbf{0.088} & 0.128 & \textbf{0.180} & 0.039 & 0.053 & \textbf{0.041} & \textbf{0.046} & \textbf{0.057} & \textbf{0.111}  \\ \hline
    \end{tabular}
    \label{tab:euroc_rmse_table}
    \vspace{-2ex}
\end{table*}

\begin{table*}[t]
\scriptsize
\centering
\caption{Average number of keyframes for 5 runs in EuRoC sequences.}
\begin{tabular}{c|c c c c c c c c c c c}
\hline
\textbf{Algorithm}  & \textbf{MH\_01} & \textbf{MH\_02} & \textbf{MH\_03} & \textbf{MH\_04} & \textbf{MH\_05} & \textbf{V1\_01} & \textbf{V1\_02} & \textbf{V1\_03} & \textbf{V2\_01} & \textbf{V2\_02} & \textbf{V2\_03}\\ \hline
default-kf                & 64.00  & 57.80 & 91.40 & 76.00 & 70.20 & 70.60  & 119.60 & 238.80 & 74.80 & 172.00 & 281.40 \\
e93-voxel & 46.00  & 46.30 & 67.80  & 58.80 & 61.80 & 52.80  & 56.20 & 120.40  & 30.80 & 63.40  & 86.80 \\ 
e95-voxel & 71.20  & 66.00 & 87.00  & 74.40 & 75.80 & 76.40  & 86.80 & 160.00  & 39.80 & 85.00  & 107.80 \\ 
e98-voxel & 154.20 & 137.20 & 181.20 & 138.60 & 143.60 & 176.80 & 177.80 & 305.20 & 84.40 & 169.00 & 203.60 \\ \hline
\end{tabular}
\label{tab:euroc_kf_comparison_BA}
\vspace{-4ex}
\end{table*}

\subsubsection{EuRoC Dataset}
We tested the multi-camera VIO pipeline on EuRoC dataset for the stereo setup.
The number of keyframes in the sliding window was set to 10.
To show the effect of the relative negative entropy for keyframe selection, we also experimented with different relative entropy thresholds.
We use the notation ``e\textbf{r}-\textbf{m}" to denote our experimental configurations, where \textbf{r} is the entropy threshold in percentage, and \textbf{m} the map representation used (\ie voxel or keyframes).
The default pipeline that is carefully tuned for stereo setups is denoted as ``default-kf''.

The median values of the absolute trajectory error in 5 runs are shown in \Tab\ref{tab:euroc_rmse_table}.
While there is no definite winner, the adapted pipeline in general performed similar or better than the default pipeline.
This can also be confirmed from the odometry errors in \Fig\ref{fig:euroc_rel_traj_error} (we select three sequences only due to the limit of space).
The adapted pipeline has lower estimate error in 10 out of 11 sequences and the entropy ratio of 98\% has the most.
Regarding the number of keyframes, it is clearly seen in \Tab\ref{tab:euroc_kf_comparison_BA} that increasing the relative entropy ratio resulted in more keyframes.
In addition, for relative entropy ratio of $95$\%, fewer keyframes were selected in general but the accuracy was still similar to the default pipeline according to \Tab\ref{tab:euroc_rmse_table}.
This indicates that the proposed method selected keyframes more effectively and introduced less redundancy than the default pipeline.

To summarize, as a generic pipeline, our method performed at least similarly good compared with a carefully tuned stereo pipeline, and our method was able to achieve similar accuracy with fewer keyframes.
More importantly, we would like to emphasize that our method has \emph{only one} parameter for keyframe selection, which makes the task of parameter tuning much easier.

\begin{table}
\scriptsize
\centering
\caption{The average number of keyframes by different keyframe selection criteria for monocular and stereo setups.}
\begin{tabular}{c|c c c c}
\hline
\textbf{Algorithm}  & \textbf{MH\_01} & \textbf{MH\_02} & \textbf{V2\_01} & \textbf{V2\_02} \\ \hline
\zz{heuristic, mono}      & 202.75  & 190.75 & 150.75 & 379.75 \\
\zz{heuristic, stereo}    & 90.00   & 117.25 & 84.75  & 204.5   \\ \hline
\zz{entropy, mono}          & 129.5   & 128.25 & 100.00 & 193.5  \\
\zz{entropy, stereo}        & 122.25  & 125.25 & 98.5   & 195     \\ 
\hline
\end{tabular}
\label{tab:euroc_kf_comparison_no_BA}
\vspace{-2ex}
\end{table}

\noindent
\textbf{Sensor Agnostic}
We also performed an experiment comparing the number of selected keyframes between monocular and stereo configurations.
We only ran the visual front-end in this case to remove the influence of the optimization backend\jc{, which caused the different keyframe numbers between \Tab\ref{tab:euroc_kf_comparison_BA} and \ref{tab:euroc_kf_comparison_no_BA}}.
The average number of keyframes on some sequences in EuRoC is shown in \Tab\ref{tab:euroc_kf_comparison_no_BA}.
The \jc{heuristic method} selected drastically different numbers of keyframes between monocular and stereo configurations \jc{because they had to be} tuned differently for these configurations.
In contrast, our entropy-based method selected very similar numbers of keyframes.
This is due to fact that our method essentially summarizes the information in the map instead of relying on camera-dependent quantities.
In particular, the stereo pair in EuRoC dataset has largely overlapping FoVs, and thus the visible areas of the environment were similar for monocular and stereo setups, leading to similar information for our keyframe selection method.

\begin{table}[]
\scriptsize
\centering
\caption{
Different trajectory error metrics from the multi-camera pipeline on the Science Park day sequence.
The first row contains the absolute RMSE of the full trajectory (\SI{547.488}{\metre})
}
\begin{tabular}{c|c c c}
\hline
\textbf{Metric}  & \textbf{F} & \textbf{FR} & \textbf{FRB} \\ \hline
Abs. Trajectory error (meter)            & 1.184  & 2.366 & 1.766\\
Rel. Trans. Percentage @ 200m   & 0.582   & 1.808 & 1.320 \\ 
Rel. Trans. Percentage @ 400m   & 0.642   & 1.07 & 0.760 \\ \hline
\end{tabular}
\label{autovision_error_analysis}
\vspace{-2ex}
\end{table}

\begin{figure}
    \centering
    \includegraphics[width=0.3\textwidth]{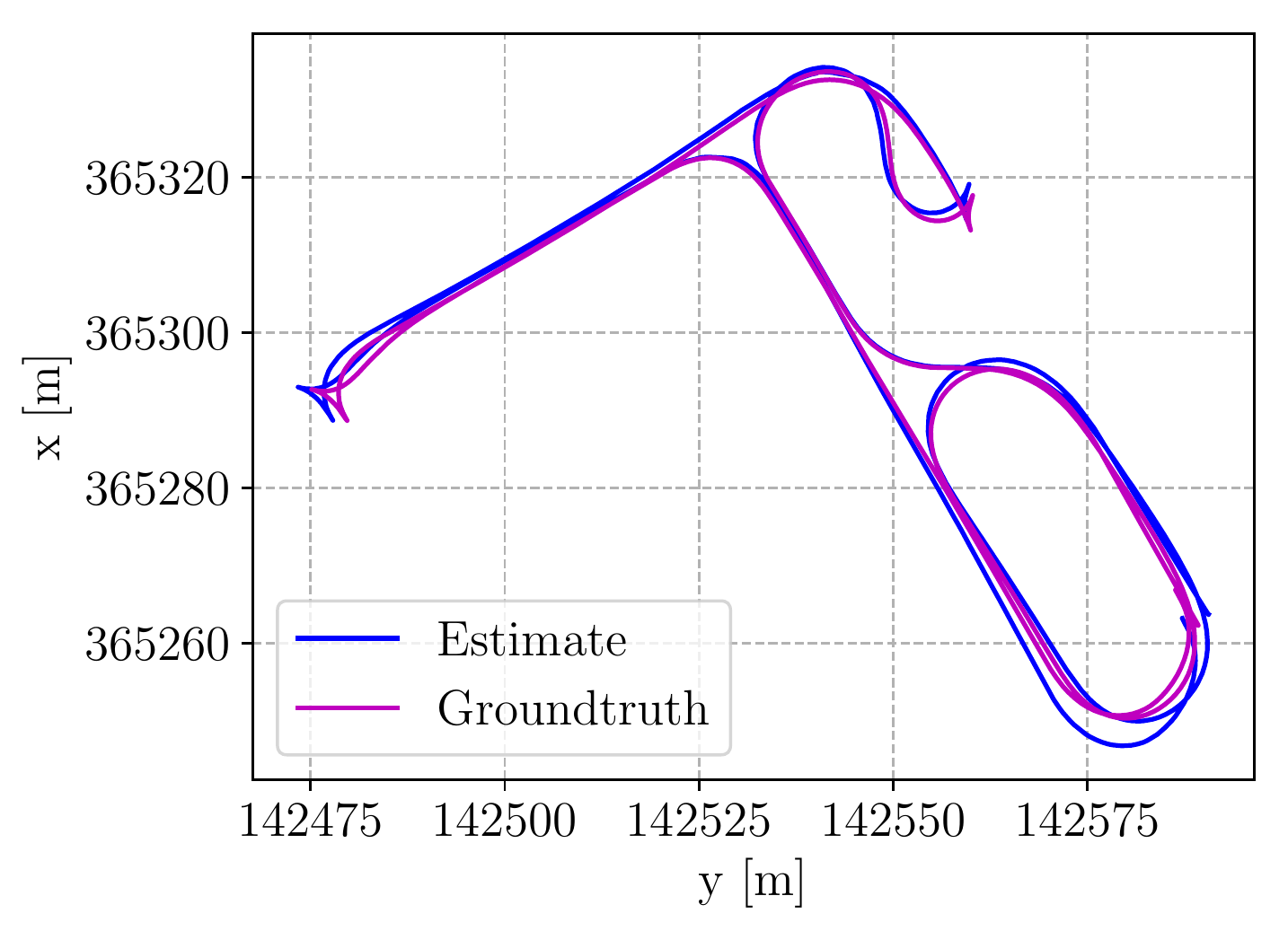}
    \caption{Top view of the estimated and groundtruth trajectory of the FRB configuration from the Science Park day sequence.}
    \vspace{-4ex}
    \label{fig:autovision_top_view}
\end{figure}

\subsubsection{AutoVision Dataset}
We evaluated our pipeline on the Science Park day sequence, which is a large-scale outdoor sequence in a autonomous driving scenario. The trajectory is 547.448 m long and the maximum speed is 3.941 m/s.
Following \cite{Liu2018IROS}, we tested our pipeline with F, FR, and FRB configurations.
The trajectory errors, computed in the same way as in \cite{Liu2018IROS}, are shown in \Tab\ref{autovision_error_analysis}, and the estimated trajectory (FRB) in \Fig\ref{fig:autovision_top_view}.
While the estimation accuracy is acceptable and proves the effectiveness of our method,  we indeed observed that the accuracy of the trajectory estimates does not necessarily increase as we add more cameras to the pipeline.
We suspect that the reason to be the inaccurate extrinsic (similar behavior can be observed in \cite{Liu2018IROS}).

\section{Conclusion}\label{sec:conclusion}
In this work, we introduced several novel designs for common building blocks in SLAM to make an adaptive system for arbitrary camera configurations.
In particular, we proposed an adaptive initialization scheme that is able to automatically find the suitable initialization method, an information-theoretic keyframe selection method that incorporates the information from all cameras elegantly and a voxel-map representation from which we can directly retrieve the landmarks in the camera FoVs.
We applied these techniques to a state-of-the-art VIO pipeline, and extensive experimental results showed that the resulting pipeline was able to adapt to various camera configurations with minimum parameter tuning. 
\IEEEtriggeratref{15}
\bibliographystyle{IEEEtran}
\bibliography{rpg_all,references}

\end{document}